\pdfoutput=1

\documentclass[11pt]{article}

\usepackage{acl}

\usepackage{times}
\usepackage{latexsym}

\usepackage[T1]{fontenc}

\usepackage[utf8]{inputenc}

\usepackage{microtype}

\usepackage{inconsolata}

\usepackage{graphicx}
\usepackage{color,soul}
\usepackage{geometry}
\usepackage{amsmath}
\usepackage{caption}
\usepackage{subcaption}
\usepackage{multirow}
\usepackage{float}
\usepackage{multirow}
\usepackage{graphicx}
\usepackage{hyperref}

\usepackage{xcolor}

\title{Document-Level Zero-Shot Relation Extraction with Entity Side Information}

\author{Mohan Raj Chanthran$^1$, Lay-Ki Soon$^{1*}$, Ong Huey Fang$^1$, and Bhawani Selvaretnam$^2$
       \\
       $^1$School of Information Technology, Monash University Malaysia\\
       \{mohanraj.chanthran, soon.layki, ong.hueyfang\}@monash.edu\\ 
       $^2$Valiantlytix\\
       {bhawani@valiantlytix.com} \\
       }

\graphicspath{ {./figures/} }

\begin{document}
\maketitle

\begin{abstract}
Document-Level Zero-Shot Relation Extraction (DocZSRE) aims to predict unseen relation labels in text documents without prior training on specific relations. Existing approaches rely on Large Language Models (LLMs) to generate synthetic data for unseen labels, which poses challenges for low-resource languages like Malaysian English. These challenges include the incorporation of local linguistic nuances and the risk of factual inaccuracies in LLM-generated data. This paper introduces Document-Level Zero-Shot Relation Extraction with Entity Side Information (DocZSRE-SI) to address limitations in the existing DocZSRE approach. The DocZSRE-SI framework leverages Entity Side Information, such as Entity Mention Descriptions and Entity Mention Hypernyms, to perform ZSRE without depending on LLM-generated synthetic data. The proposed low-complexity model achieves an average improvement of 11.6\% in the macro F1-Score compared to baseline models and existing benchmarks. By utilizing Entity Side Information, DocZSRE-SI offers a robust and efficient alternative to error-prone, LLM-based methods, demonstrating significant advancements in handling low-resource languages and linguistic diversity in relation extraction tasks. This research provides a scalable and reliable solution for ZSRE, particularly in contexts like Malaysian English news articles, where traditional LLM-based approaches fall short.
\end{abstract}

\section{Introduction}
\label{sec:introduction}

Relation Extraction (RE) is a crucial NLP task that identifies the relation between entities in text. Document-Level Relation Extraction (DocRE) takes this further by capturing relations across sentences. Most RE models rely on supervised learning, which requires large amounts of labelled data and is limited to predicting predefined relations. Labelling data is expensive and time-consuming, especially for news articles where new relations constantly emerge. Open Relation Extraction (ORE) tries to address this by identifying relation phrases without predefined labels, but it often produces redundant or overly specific results, making standardization and interpretation difficult. Zero-Shot Relation Extraction (ZSRE) has emerged as a solution to overcome the limitations of supervised and ORE approaches. While significant progress has been made in sentence-level ZSRE, document-level ZSRE remains largely unexplored. Currently, only one notable approach exists for document-level ZSRE \citep{10.1145/3589334.3645678}, which generates synthetic data for unseen relations and fine-tunes a language model on seen relations. However, this method is complex, resource-intensive, and relies heavily on LLMs like ChatGPT, which struggle with low-resource languages such as Malaysian English \citep{chanthran2023chatgpt}. These limitations highlight the need for more robust and scalable solutions.

To address these gaps, we propose Document-Level Zero-Shot Relation Extraction with Entity Side Information (DocZSRE-SI). This framework tackles the challenges of document-level ZSRE by focusing on Entity Side Information, including Entity Mention Descriptions, Entity Mention Hypernyms, and Entity Types. Instead of processing entire documents, DocZSRE-SI concentrates on the Entity Mention Descriptions relevant to the entity pairs being evaluated. This approach provides a concise and meaningful representation of entities, improving efficiency and accuracy by focusing on essential context and reducing noise from long documents. Our evaluation shows that DocZSRE-SI performs well not only in Malaysian English but also in Standard English. The key contributions of this paper are as follows:
\begin{enumerate}
    \item Introduction of DocZSRE-SI: We propose Document-Level Zero-Shot Relation Extraction framework that leverages Entity Side Information to enhance the prediction of unseen relations within a document. The code of this framework is published in \url{https://github.com/mohanraj-nlp/DocZSRE-SI}. 
    \item Efficient Context Utilization: Instead of processing entire documents, DocZSRE-SI focuses on relevant Entity Mention Descriptions, reducing noise and improving efficiency while maintaining high accuracy in predicting unseen relations.
    \item Incorporation of Entity Side Information: Our approach integrates Entity Mention Hypernym, and Entity Type to provide a richer semantic representation, improving the model’s ability to infer relations in a zero-shot setting.
\end{enumerate}

This paper is structured as follows: Section \ref{sec:related_work} reviews existing approaches for sentence-level and document-level ZSRE. Section \ref{sec:methodology} presents the DocZSRE-SI framework, which consists of two modules: Building Entity Side Information and Zero-Shot Relation Extraction. In Sections \ref{sec:experiments} and \ref{sec:resultsanddiscussion}, we describe the experimental setup and discuss the results. Finally, Section \ref{sec:conclusion} concludes the paper and outlines directions for future work.

\section{Related Work}
\label{sec:related_work}

ZSRE extracts relations between entities without task-specific labelled data, using pre-trained language models (PLMs), knowledge graphs, or prompt-based methods to generalize to unseen relations. While most ZSRE approaches focus on Sentence-Level (intra-sentential) Relation Extraction (Section \ref{ssec:related_work_sentence_level_zsre}), only one notable work addresses Document-Level (inter-sentential) Relation Extraction (Section \ref{ssec:related_work_document_level_zsre}). This section reviews key ZSRE contributions and is grouped by techniques.

\subsection{Sentence-Level ZSRE}
\label{ssec:related_work_sentence_level_zsre}
Sentence-Level Zero-Shot Relation Extraction (ZSRE) has been explored through various methodologies, including reformulating it as Reading Comprehension and Textual Entailment tasks, leveraging Prompt-based Learning with external knowledge, and advancing Representation Learning with matching techniques. Early works framed ZSRE as reading comprehension \citep{levy-etal-2017-zero} or textual entailment tasks \citep{obamuyide-vlachos-2018-zero}, with later improvements like entailment templates \citep{rahimi-surdeanu-2023-improving}. Prompt-based methods, such as RelationPrompt \citep{chia-etal-2022-relationprompt} and ZS-SKA \citep{gong-eldardiry-2024-prompt}, combined prompts with external knowledge, achieving strong results on datasets like FewRel \citep{han-etal-2018-fewrel}, and Wiki-ZSL \citep{chen-li-2021-zs} but struggling with generalization to unseen relations. Representation learning approaches, including ZSLRC \citep{gong2020zeroshot}, ZS-BERT \citep{chen-li-2021-zs}, and RE-Matching \citep{zhao-etal-2023-matching}, focused on fine-grained matching and improved zero-shot classification. Additionally, Weak Supervision and Template Infilling have emerged as innovative strategies to reduce reliance on annotated data.

\subsection{Document-Level ZSRE}
\label{ssec:related_work_document_level_zsre}
The literature review highlights limited work on Document-Level Zero-Shot Relation Extraction (ZSRE). One notable contribution is by \citep{10.1145/3589334.3645678}, which proposes generating synthetic data using ChatGPT to handle unseen relation labels. The approach introduces a Chain-of-Retrieval (CoR) prompt to guide the generation of sentences corresponding to relation triplets and incorporate a Consistency-Guided Knowledge Denoising Strategy to enhance the quality of synthetic data. Experiments show the approach achieves 41.3 ± 8.9 and 41.5 ± 8.7 on Re-DocRED and DocRED test sets, outperforming baselines and demonstrating its effectiveness in generating high-quality relational data without extensive human annotation. Despite this advancement, document-level ZSRE remains underexplored, calling for further research.

\section{Methodology}
\label{sec:methodology}

\begin{figure*}[!ht]
    \centering
    \includegraphics[height=\textheight,width=\textwidth,keepaspectratio]{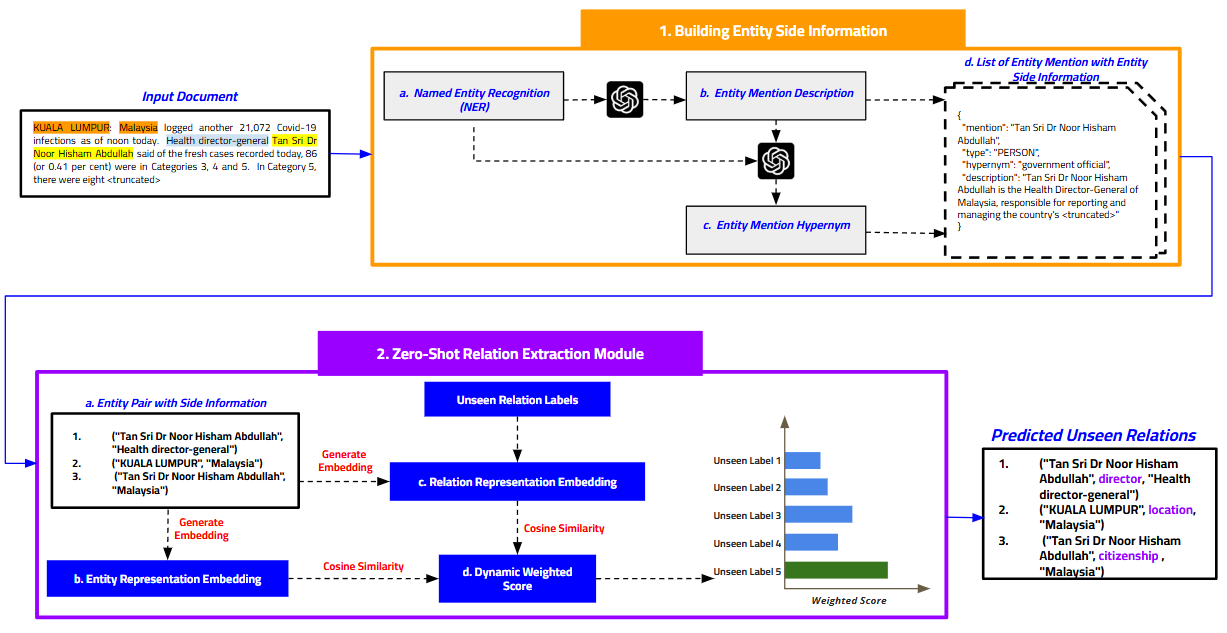}
    \caption{High-Level architecture of DocZSRE-SI Framework}
    \label{fig:high_level_architecture_desire}
\end{figure*}

Figure \ref{fig:high_level_architecture_desire} provides an overview of DocZSRE-SI, which consists of two key components. First, the Building Entity Side Information Module processes input document to extract additional details of Entity Mention like Entity Types, Entity Mention Descriptions, and Entity Mention Hypernyms (details Section \ref{ssec:building_entity_side_information}). Second, the Zero-Shot Relation Extraction Module analyzes each entity pair to identify the best unseen relation label. This is done by leveraging Entity Side Information and calculating a Dynamic Weighted Score for each label. The label with the highest score is selected as the correct relation (details in Section \ref{ssec:unseen_relation_inference_module}).

\subsection{Building Entity Side Information}
\label{ssec:building_entity_side_information}
Building Entity Side Information module is a key element of the DocZSRE-SI framework. This component is designed to enrich information about entities based on input documents. The collected side information includes Entity Mention Type, Entity Mention Description (Section \ref{sssec:entity_mention_description_gen}), and Entity Mention Hypernym (Section \ref{sssec:entity_mention_hypernym_gen}).

\subsubsection{Entity Mention Description Generator}
\label{sssec:entity_mention_description_gen}
Entity Mention Description extends traditional NER by providing rich, contextual details about an entity. Instead of simply labelling it as a \textit{PERSON}, \textit{ORGANIZATION}, or \textit{LOCATION}, it generates a detailed textual description that captures the surrounding text, semantic context, and implied attributes while also pulling information from multiple sentences or paragraphs for a more comprehensive understanding. This approach is used instead of processing the full document because it focuses on the most relevant context for the entity pair, reducing noise and improving efficiency. These descriptions are generated using gpt-4o-mini \citep{openai_gpt4}, chosen after testing various LLMs for quality. An example of a generated description can be found in Appendix \ref{sec:entity_mention_description_appendix}.

\subsubsection{Entity Mention Hypernym Generator}
\label{sssec:entity_mention_hypernym_gen}
Hypernyms represent the broader category an entity belongs to, providing a higher-level understanding of its classification. Hypernyms can help differentiate entities with the same entity type but different roles. For example, consider these two entity mentions, \textit{Maybank Sdn Bhd} and \textit{Khairussaleh Ramli}. For context, \textit{Khairussaleh Ramli} is the CEO of \textit{Maybank Sdn Bhd}. The entity type of \textit{Maybank Sdn Bhd} is \textit{ORGANIZATION}, and \textit{Khairussaleh Ramli} is \textit{PERSON}, but these labels alone do not provide sufficient context about the entities. By incorporating hypernyms, it becomes more informative that \textit{Maybank Sdn Bhd} is a \textit{banking institution}, and \textit{Khairussaleh Ramli} is a \textit{business executive}. Previous work \citep{gong2020zeroshot} showed that hypernyms improve sentence-level ZSRE. These hypernyms are generated by combining entity mention, entity mention type, and entity mention description using gpt-4o-mini.

\subsection{Zero-Shot Relation Extraction Module}
\label{ssec:unseen_relation_inference_module}
The Zero-Shot Relation Extraction Module is responsible for identifying unseen relation labels. The module includes components like Entity Side Information Embedding (Section \ref{sssec:entity_side_information_embedding}), Relation Representation Embedding (Section \ref{sssec:relation_information_embedding}), and Dynamic Weighted Score (Section \ref{sssec:dynamic_weighted_score}).

\subsubsection{Entity Side Information Embedding}
\label{sssec:entity_side_information_embedding}
The Entity Side Information Embedding component combines the different entity-side features and produces embeddings that are then used in subsequent stages of the framework for ZSRE. These embeddings are derived from different aspects of the entity side information:
\begin{enumerate}
    \item \textbf{Combined Description Embedding}: Merging the descriptions of head and tail entities provides richer context, helping the model understand their roles. However, descriptions alone may not fully capture the relation between entities.
    \item \textbf{Entity Hypernym Embeddings}: Hypernyms offer higher-level categories for entities, improving generalization to unseen relations by helping the model recognize broader patterns.
    \item \textbf{Entity Type Embeddings}: Entity types help differentiate entities by providing basic category information. Combining entity types with hypernyms balances general and specific details, improving relation prediction.
    \item \textbf{Role-Based Embeddings}: Roles of entity clarify how entities function within a relation by distinguishing between subjects and objects. This prevents the model from misinterpreting entity relationships. Prompt templates for role-based embeddings are defined in Appendix \ref{sec:prompt_role_based_embedding}
    \item \textbf{Context Embedding}: This component generates an embedding to capture the contextual relation between two entities. It is designed to calculate the cosine similarity with the relation label embedding, aiding in identifying the most likely relation. The prompt used for context embedding is defined in Appendix \ref{sec:prompt_context_embedding}. The prompt conveys the idea of a connection between two things, where the {head\_hypernym} and {tail\_hypernym} represent the entities involved. This prompt was included as additional information based on our observation that entity types are often used as arguments when predicting relations.
\end{enumerate}
In this work, we use a pre-trained BERT model \citep{devlin-etal-2019-bert}, specifically \texttt{bert-base-uncased} \footnote{\url{https://huggingface.co/google-bert/bert-base-uncased
}}, to generate the embeddings. Entity descriptions, types, and hypernyms provide useful context but may not fully capture entity relationships. Descriptions can be vague, hypernyms may overlook unique interactions, and entity types alone lack depth. Combining these with role-based embeddings enhances the model’s understanding, thereby improving accuracy in predicting unseen relations.

\subsubsection{Relation Label Embedding }
\label{sssec:relation_information_embedding}
Relation Label Embedding converts relation labels into dense vectors, representing unseen relations. We encode each relation label using \texttt{bert-base-uncased}\footnote{\url{https://huggingface.co/google-bert/bert-base-uncased
}}, following the same encoding strategy used for entity representations. Cosine similarity with Entity Side Information Embeddings helps assess how well an entity pair aligns with a relation. Calculating cosine similarity with Entity Side Information Embeddings helps measure how well the current entity pair aligns with the specific relation. Section \ref{sssec:calculating_cosine_similarity} explains this calculation in detail.

\subsubsection{Calculating Cosine Similarity}
\label{sssec:calculating_cosine_similarity}
\begin{figure*}[t] 
\centering
\begin{equation}
\text{Role-Based Score} = \frac{(\text{Head Role-Based Embedding} + \text{Tail Role-Based Embedding})}{2}
\label{eq:head_and_tail_role_emb}
\end{equation}
\end{figure*}
    
Cosine similarity is used to measure the similarity between the embeddings discussed in Sections \ref{sssec:entity_side_information_embedding} and \ref{sssec:relation_information_embedding}, allowing the framework to assess how closely the Entity Side Information Embedding matches the target Relation Label Embedding. Cosine similarity measures the angle between two vectors, with values ranging from -1 (completely dissimilar) to +1 (completely similar). The framework computes multiple similarity measures:
\begin{enumerate}
    \item \textbf{Description Similarity Score}: Measures alignment between Relation Label Embedding and Combined Description Embedding.
    \item \textbf{Entity Hypernym Similarities Score}: Separate similarity scores are calculated between the Relation Label Embedding and the Entity Hypernym Embeddings of the head and tail entities.
    \item \textbf{Entity Type Similarities Score}: Separate similarity scores are calculated between the Relation Label Embedding and the Entity Type Embeddings of the head and tail entities. 
    \item \textbf{Role-Based Similarities Score}: Equation \ref{eq:head_and_tail_role_emb} shows how the role-based similarity score are calculated. Separate scores are calculated between role-based embeddings and relation label context. Role-based embeddings clarify entity roles (subject/object) and resolve ambiguities from types or hypernyms. With two embeddings (head and tail entities), their average is calculated to prevent one entity's information (Type and Hypernym) from heavily influencing the result.
    \item \textbf{Context Similarity Score}: Evaluates how well the relation between two entity types matches the target label. For instance, a relation between \textit{\{person\}} and \textit{\{educational institution\}} might align with labels like \textit{educated\_at, place\_of\_birth}, adding context for better predictions.
    \item \textbf{Consistency-Based Confidence Weightage}: The confidence score combines the mean and consistency of six similarity measures. Consistency is measured by the standard deviation (lower = more agreement), while mean similarity reflects strength. Averaging these ensures higher confidence when scores are both strong (High Mean) and stable (Low Deviation), making predictions more reliable.
\end{enumerate}
Different similarity measures help the model better understand the relation between entities, especially with unseen relation labels. However, simply summing these scores is not enough for accurate predictions. To improve accuracy, more weight is given to the most important scores, ensuring a stronger influence on the final result (see Section \ref{sssec:dynamic_weighted_score}).

\subsubsection{Dynamic Weighted Score}
\label{sssec:dynamic_weighted_score}
Dynamic Weighted Score enhances relation prediction by assigning different importance levels to similarity scores. It prioritizes key features, like entity descriptions, over others (e.g., entity types or hypernyms), ensuring better entity-relation alignment. Weightage will be assigned to each similarity score based on its relevance.
\begin{multline}
\text{Dynamic Weighted Score} = \Big( \\0.4 \times \text{Description Similarity (a, r)} \\ + 0.1 \times \text{Entity Hypernym Similarity (b, r)} \\ +
0.1 \times \text{Entity Hypernym Similarity (c, r)} \\ + 0.1 \times \text{Entity Type Similarity (d, r)} \\ + 
0.1 \times \text{Entity Type Similarity (e, r)} \\ + 0.1 \times \text{Role-Based Similarities Score (f,g,r)} \\ +
0.1 \times \text{Context Similarity Score} \Big) \\ \times \text{Consistency-Based Confidence Weightage}
\end{multline}
\label{eq:weighted_score_formula}
where:
\[
\begin{aligned}
a & = \text{Combined Description Embedding} \\
b & = \text{Head Entity Hypernym Embedding} \\
c & = \text{Tail Entity Hypernym Embedding} \\
d & = \text{Head Entity Type Embedding} \\
e & = \text{Tail Entity Type Embedding} \\
f & = \text{Head Role-Based Embedding} \\
g & = \text{Tail Role-Based Embedding} \\
h & = \text{Tail Entity Type Embedding} \\
r & = \text{Relation Label Embedding}
\end{aligned}
\]

In Equation \ref{eq:weighted_score_formula}, description similarity is weighted highest (0.4), reflecting its importance in capturing entity interactions. Descriptions provide rich contextual details, making them more impactful than other features. We tested weights (0.2, 0.4, 0.6) and chose 0.4 for optimal balance, where lower weights reduced its influence while higher weights overshadowed other features. This ensures description similarity remains significant without diminishing other contributions. The unseen relation label with the highest score is selected, improving prediction accuracy by prioritising contextually relevant features.

\section{Experiments}
\label{sec:experiments}

\subsection{Evaluation Metrics}
\label{ssec:eval_metrics}

ZSRE evaluation follows prior works \citep{gong-eldardiry-2024-prompt, chen-li-2021-zs, chia-etal-2022-relationprompt, wang-etal-2022-rcl, zhao-etal-2023-matching, kim-etal-2023-zero, 10.1145/3589334.3645678}, where unseen relations are randomly chosen from the dataset. Unseen relation sets of sizes \( n \in \{5, 10, 15\} \) are used, with three random samples for each size to ensure results aren’t biased by specific relation choices. This random sampling tests the model’s generalization across different scenarios. Macro F1-Score is the primary metric, calculated for each run and averaged, with variance reported to measure consistency. Low variance indicates stable and reliable performance, while high variance suggests sensitivity to specific data samples. This setup ensures a fair and robust evaluation.

\subsection{Dataset and Benchmarking}
\label{ssec:dataset_and_benchmarking}
This research uses DocRE datasets, including MEN-Dataset \citep{chanthran-etal-2024-malaysian}, DocRED \citep{yao-etal-2019-docred}, and RE-DocRED \citep{tan-etal-2022-revisiting}. We split our experiments into two parts. The first part consists of an ablation study conducted on the MEN dataset, RE-DocRED, and a subset of the DocRED dataset, where only 20\% of the documents (21,577 documents) are used due to computational constraints. The second part evaluates the model using the full development and test sets of DocRED and RE-DocRED, respectively. Unseen relation labels will be randomly selected for fair evaluation. A baseline will be established using only Entity Mention Descriptions, excluding features such as Entity Type, Hypernym, or Dynamic Weighted Score. The proposed framework will be compared with existing document-level ZSRE methods, particularly \citep{10.1145/3589334.3645678}.

\subsection{Experiments Planned}
\label{ssec:planned_experiment}
To evaluate DocZSRE-SI, we conduct two key experiments. First, an ablation study examines the impact of several key features, including Entity Mention Descriptions, Entity Mention Types, Entity Mention Hypernyms, and a Weighted Dynamic Scoring mechanism. This helps us to better understand how each feature contributes to the overall performance of our approach and identify the optimal combination for improving prediction accuracy. The results of this experiment are presented in Table \ref{tab:ablation_study_full_unseen_relation_inference}. As this is a ZSRE, we conduct experiments on the full MEN dataset, the RE-DocRED dataset, and 20\% of the DocRED dataset. Second, we compare our approach with GenRDK, a method proposed by \citep{10.1145/3589334.3645678} for predicting unseen relations. Performance is evaluated on the dev and test sets of DocRED and RE-DocRED. The unseen relation labels are randomly selected to ensure fairness. This comparison offers insights into the effectiveness of our framework in comparison to existing Document-Level ZSRE methods.

\section{Results and Discussion}
\label{sec:resultsanddiscussion}

\subsection{Ablation Study}
\label{ssec:ablationstudy}

    

\begin{table}[]
\centering
\resizebox{\columnwidth}{!}{%
\begin{tabular}{|cc|c|c|c|}
\hline
\multicolumn{2}{|c|}{} &
  RE-DocRED &
  DocRED &
  MEN-Dataset \\ \hline
\multicolumn{1}{|c|}{} &
  {\color[HTML]{000000} \begin{tabular}[c]{@{}c@{}}Only Entity Mention  \\ Description (Baseline)\end{tabular}} &
  \begin{tabular}[c]{@{}c@{}}28.14\\ ±\\ 5.43\end{tabular} &
  \begin{tabular}[c]{@{}c@{}}36.34\\ ±\\ 5.43\end{tabular} &
  \begin{tabular}[c]{@{}c@{}}27.65\\ ±\\ 24.46\end{tabular} \\ \cline{2-5} 
\multicolumn{1}{|c|}{} &
  \begin{tabular}[c]{@{}c@{}}Entity Mention \\ Description + Entity \\ Mention Hypernym\end{tabular} &
  \begin{tabular}[c]{@{}c@{}}42.34\\ ±\\ 2.39\end{tabular} &
  \begin{tabular}[c]{@{}c@{}}42.23\\ ±\\ 2.39\end{tabular} &
  \begin{tabular}[c]{@{}c@{}}33.96\\ ±\\ 4.88\end{tabular} \\ \cline{2-5} 
\multicolumn{1}{|c|}{} &
  \begin{tabular}[c]{@{}c@{}}Entity Mention \\ Description + Entity \\ Mention Type\end{tabular} &
  \begin{tabular}[c]{@{}c@{}}35.12\\ ±\\ 5.65\end{tabular} &
  \begin{tabular}[c]{@{}c@{}}31.41\\ ±\\ 5.65\end{tabular} &
  \begin{tabular}[c]{@{}c@{}}28.33\\ ±\\ 15.76\end{tabular} \\ \cline{2-5} 
\multicolumn{1}{|c|}{\multirow{-5}{*}{5}} &
  \begin{tabular}[c]{@{}c@{}}Entity Mention \\ Description + Entity \\ Mention Hypernym + \\ Entity Mention Type\end{tabular} &
  \begin{tabular}[c]{@{}c@{}}38.67\\ ±\\ 8.1\end{tabular} &
  \begin{tabular}[c]{@{}c@{}}36.44\\ ±\\ 8.15\end{tabular} &
  \begin{tabular}[c]{@{}c@{}}39.79\\ ±\\ 10.66\end{tabular} \\ \cline{2-5} 
\multicolumn{1}{|c|}{} &
  \begin{tabular}[c]{@{}c@{}}Entity Mention \\ Description + Entity \\ Mention Hypernym + \\ Entity Mention Type + \\ Dynamic Weighted \\ Score (Proposed \\ Approach)\end{tabular} &
  \begin{tabular}[c]{@{}c@{}}50.05 \\ ±\\ 8.37\end{tabular} &
  \begin{tabular}[c]{@{}c@{}}48.83\\ ±\\ 7.57\end{tabular} &
  \begin{tabular}[c]{@{}c@{}}40.25\\ ±\\ 7.53\end{tabular} \\ \hline
\multicolumn{1}{|c|}{} &
  \begin{tabular}[c]{@{}c@{}}Only Entity Mention  \\ Description (Baseline)\end{tabular} &
  \begin{tabular}[c]{@{}c@{}}20.47\\ ±\\ 5.81\end{tabular} &
  \begin{tabular}[c]{@{}c@{}}22.36\\ ±\\ 5.81\end{tabular} &
  \begin{tabular}[c]{@{}c@{}}19.89\\ ±\\ 18.19\end{tabular} \\ \cline{2-5} 
\multicolumn{1}{|c|}{} &
  \begin{tabular}[c]{@{}c@{}}Entity Mention \\ Description + Entity \\ Mention Hypernym\end{tabular} &
  \begin{tabular}[c]{@{}c@{}}39.89\\ ±\\ 12.79\end{tabular} &
  \begin{tabular}[c]{@{}c@{}}37.7\\ ±\\ 12.79\end{tabular} &
  \begin{tabular}[c]{@{}c@{}}30.75\\ ±\\ 10.95\end{tabular} \\ \cline{2-5} 
\multicolumn{1}{|c|}{} &
  \begin{tabular}[c]{@{}c@{}}Entity Mention \\ Description + Entity \\ Mention Type\end{tabular} &
  \begin{tabular}[c]{@{}c@{}}25.82\\ ±\\ 6.35\end{tabular} &
  \begin{tabular}[c]{@{}c@{}}22.67\\ ±\\ 6.35\end{tabular} &
  \begin{tabular}[c]{@{}c@{}}23.49\\ ±\\ 6.52\end{tabular} \\ \cline{2-5} 
\multicolumn{1}{|c|}{\multirow{-5}{*}{10}} &
  \begin{tabular}[c]{@{}c@{}}Entity Mention \\ Description + Entity \\ Mention Hypernym + \\ Entity Mention Type\end{tabular} &
  \begin{tabular}[c]{@{}c@{}}35.83\\ ±\\ 14.2\end{tabular} &
  \begin{tabular}[c]{@{}c@{}}33.59\\ ±\\ 14.1\end{tabular} &
  \begin{tabular}[c]{@{}c@{}}26.13\\ ±\\ 9.32\end{tabular} \\ \cline{2-5} 
\multicolumn{1}{|c|}{} &
  \begin{tabular}[c]{@{}c@{}}Entity Mention \\ Description + Entity \\ Mention Hypernym + \\ Entity Mention Type + \\ Dynamic Weighted \\ Score (Proposed \\ Approach)\end{tabular} &
  \begin{tabular}[c]{@{}c@{}}44.98\\ ±\\ 6.78\end{tabular} &
  \begin{tabular}[c]{@{}c@{}}43.43\\ ±\\ 8.64\end{tabular} &
  \begin{tabular}[c]{@{}c@{}}36.02\\ ±\\ 8.22\end{tabular} \\ \hline
\multicolumn{1}{|c|}{} &
  \begin{tabular}[c]{@{}c@{}}Only Entity Mention  \\ Description (Baseline)\end{tabular} &
  \begin{tabular}[c]{@{}c@{}}8.56\\ ± \\ 5.05\end{tabular} &
  \begin{tabular}[c]{@{}c@{}}8.4\\ ±\\ 5.05\end{tabular} &
  \begin{tabular}[c]{@{}c@{}}16.39\\ ±\\ 4.84\end{tabular} \\ \cline{2-5} 
\multicolumn{1}{|c|}{} &
  \begin{tabular}[c]{@{}c@{}}Entity Mention \\ Description + Entity \\ Mention Hypernym\end{tabular} &
  \begin{tabular}[c]{@{}c@{}}31.66\\ ±\\ 9.77\end{tabular} &
  \begin{tabular}[c]{@{}c@{}}31.02\\ ±\\ 9.77\end{tabular} &
  \begin{tabular}[c]{@{}c@{}}22.74\\ ±\\ 4\end{tabular} \\ \cline{2-5} 
\multicolumn{1}{|c|}{} &
  \begin{tabular}[c]{@{}c@{}}Entity Mention \\ Description + Entity \\ Mention Type\end{tabular} &
  \begin{tabular}[c]{@{}c@{}}16.78\\ ±\\ 3.88\end{tabular} &
  \begin{tabular}[c]{@{}c@{}}15.41\\ ±\\ 3.88\end{tabular} &
  \begin{tabular}[c]{@{}c@{}}22.64\\ ±\\ 6.22\end{tabular} \\ \cline{2-5} 
\multicolumn{1}{|c|}{\multirow{-5}{*}{15}} &
  \begin{tabular}[c]{@{}c@{}}Entity Mention \\ Description + Entity \\ Mention Hypernym + \\ Entity Mention Type\end{tabular} &
  \begin{tabular}[c]{@{}c@{}}31.67\\ ± \\ 4.25\end{tabular} &
  \begin{tabular}[c]{@{}c@{}}29.76\\ ±\\ 4.11\end{tabular} &
  \begin{tabular}[c]{@{}c@{}}23.11\\ ±\\ 2.07\end{tabular} \\ \cline{2-5} 
\multicolumn{1}{|c|}{} &
  \begin{tabular}[c]{@{}c@{}}Entity Mention \\ Description + Entity \\ Mention Hypernym + \\ Entity Mention Type + \\ Dynamic Weighted \\ Score (Proposed \\ Approach)\end{tabular} &
  \begin{tabular}[c]{@{}c@{}}33.91\\ ±\\ 4.57\end{tabular} &
  \begin{tabular}[c]{@{}c@{}}32.55\\ ±\\ 4.57\end{tabular} &
  \begin{tabular}[c]{@{}c@{}}24.28\\ ±\\ 5.1\end{tabular} \\ \hline
\end{tabular}%
}
\caption{The complete result of the Ablation Study was conducted for the Zero-Shot Relation Extraction Module for DocRED, RE-DocRED, and MEN-Dataset. Reported in this result are F1-Score with the Variance.}
\label{tab:ablation_study_full_unseen_relation_inference}
\end{table}
Various Entity Side Information was introduced to enhance the proposed methodology. An ablation study was conducted on three different datasets to evaluate its effectiveness. The study compared five approaches for predicting unseen relations, using the macro F1-Score as the primary metric. Table \ref{tab:ablation_study_full_unseen_relation_inference} shows the performance of each approach based on Macro F1-Score together with Variance. Further analysis has been conducted to compare the performance of various approaches and the impact of Entity Side Information and Dynamic Weighted Scores. 

\paragraph{Performance Comparison Across Different Approaches} \mbox{}\\
\label{para:performance_comparison_across_approaches}
The proposed approach, combining Entity Mention Hypernym, Entity Mention Type, and Dynamic Weighted Score, achieved the highest macro F1-Score across three datasets (RE-DocRED, DocRED, MEN-Dataset). The proposed approach outperformed the Baseline (using only Entity Mention Descriptions) by 164.56\% (RE-DocRED), 83.21\% (DocRED), and 35.95\% (MEN-Dataset). It correctly predicted 90\% of unseen labels, with DocRED showing the highest accuracy improvements (15-30\%) and MEN-Dataset the lowest, likely due to its smaller size and fewer instances. Low variance (e.g., 6.57 for RE-DocRED) indicated robustness across unseen relation groups, demonstrating consistent and accurate results.

\paragraph{Impact of Different Entity Side Information} \mbox{}\\
\label{para:impact_of_different_entity_side_information}
Each type of Entity Side Information contributes differently to performance. The Baseline uses only Entity Mention Descriptions, replacing full documents with combined descriptions. The ablation study evaluates the impact of adding Entity Mention Type and Entity Mention Hypernym, individually and combined. 

Results in Table \ref{tab:ablation_study_full_unseen_relation_inference} show that combining Entity Mention Descriptions with Hypernyms consistently outperforms combining descriptions with Types, with performance gaps of 45.63\% (RE-DocRED), 52.43\% (DocRED), and 18.13\% (MEN-Dataset). This highlights hypernyms' stronger contribution, providing broader semantic understanding and aiding generalization across unseen relations. In contrast, Entity Mention Types, while useful, are limited by their high-level categorical nature. Interestingly, combining all three (descriptions, types, and hypernyms) did not improve performance, as too many similarity scores introduced noise. This led to the development of a Dynamic Weighting Mechanism (Equation \ref{eq:weighted_score_formula}), which prioritizes significant information sources and boosts prediction confidence by assigning appropriate weights to similarity measures.

\begin{table*}[]
\centering
\resizebox{\textwidth}{!}{%
\begin{tabular}{|l|cccccc|cccccc|cccc|}
\hline
n &
  \multicolumn{6}{c|}{5} &
  \multicolumn{6}{c|}{10} &
  \multicolumn{4}{c|}{15} \\ \hline
Approach &
  \multicolumn{2}{c|}{GenRDK} &
  \multicolumn{2}{c|}{\begin{tabular}[c]{@{}c@{}}Best \\ Approach**\end{tabular}} &
  \multicolumn{2}{c|}{\begin{tabular}[c]{@{}c@{}}\%\\ Improvement\end{tabular}} &
  \multicolumn{2}{c|}{GenRDK} &
  \multicolumn{2}{c|}{\begin{tabular}[c]{@{}c@{}}Best \\ Approach**\end{tabular}} &
  \multicolumn{2}{c|}{\begin{tabular}[c]{@{}c@{}}\%\\ Improvement\end{tabular}} &
  \multicolumn{2}{c|}{GenRDK} &
  \multicolumn{2}{c|}{\begin{tabular}[c]{@{}c@{}}Best \\ Approach**\end{tabular}} \\ \hline
Dataset &
  \multicolumn{1}{c|}{dev} &
  \multicolumn{1}{c|}{test} &
  \multicolumn{1}{c|}{dev} &
  \multicolumn{1}{c|}{test} &
  \multicolumn{1}{c|}{dev} &
  test &
  \multicolumn{1}{c|}{dev} &
  \multicolumn{1}{c|}{test} &
  \multicolumn{1}{c|}{dev} &
  \multicolumn{1}{c|}{test} &
  \multicolumn{1}{c|}{dev} &
  test &
  \multicolumn{1}{c|}{dev} &
  \multicolumn{1}{c|}{test} &
  \multicolumn{1}{c|}{dev} &
  test \\ \hline
Re-DocRED &
  \multicolumn{1}{c|}{\begin{tabular}[c]{@{}c@{}}39.9 \\ ± \\ 10.9\end{tabular}} &
  \multicolumn{1}{c|}{\begin{tabular}[c]{@{}c@{}}41.3 \\ ± \\ 8.9\end{tabular}} &
  \multicolumn{1}{c|}{\textbf{\begin{tabular}[c]{@{}c@{}}50.95\\ ± \\ 10.2\end{tabular}}} &
  \multicolumn{1}{c|}{\textbf{\begin{tabular}[c]{@{}c@{}}47.81\\ ±\\ 9.78\end{tabular}}} &
  \multicolumn{1}{l|}{+27.69} &
  \multicolumn{1}{l|}{+15.76} &
  \multicolumn{1}{c|}{\begin{tabular}[c]{@{}c@{}}30.6 \\ ± \\ 3.6\end{tabular}} &
  \multicolumn{1}{c|}{\begin{tabular}[c]{@{}c@{}}30.1 \\ ± \\ 4.2\end{tabular}} &
  \multicolumn{1}{c|}{\textbf{\begin{tabular}[c]{@{}c@{}}47.7\\ ± \\ 12.71\end{tabular}}} &
  \multicolumn{1}{c|}{\textbf{\begin{tabular}[c]{@{}c@{}}43.06\\ ±\\ 15.5\end{tabular}}} &
  \multicolumn{1}{l|}{+55.8} &
  \multicolumn{1}{l|}{+43.05} &
  \multicolumn{1}{c|}{-} &
  \multicolumn{1}{c|}{-} &
  \multicolumn{1}{c|}{\textbf{\begin{tabular}[c]{@{}c@{}}30.87\\ ± \\ 4.4\end{tabular}}} &
  \textbf{\begin{tabular}[c]{@{}c@{}}33.88\\ ±\\ 5.55\end{tabular}} \\ \hline
\multicolumn{1}{|c|}{DocRED} &
  \multicolumn{1}{c|}{\begin{tabular}[c]{@{}c@{}}42.5 \\ ± \\ 10.6\end{tabular}} &
  \multicolumn{1}{c|}{\begin{tabular}[c]{@{}c@{}}41.5 \\ ± \\ 8.7\end{tabular}} &
  \multicolumn{1}{c|}{\textbf{\begin{tabular}[c]{@{}c@{}}45.12\\ ± \\ 7.31\end{tabular}}} &
  \multicolumn{1}{c|}{\textbf{-}} &
  \multicolumn{1}{c|}{+6.34} &
  - &
  \multicolumn{1}{c|}{\begin{tabular}[c]{@{}c@{}}33.7 \\ ± \\ 4.0\end{tabular}} &
  \multicolumn{1}{c|}{\begin{tabular}[c]{@{}c@{}}31.4 \\ ± \\ 4.6\end{tabular}} &
  \multicolumn{1}{c|}{\textbf{\begin{tabular}[c]{@{}c@{}}40.51\\ ± \\ 12.65\end{tabular}}} &
  \multicolumn{1}{c|}{\textbf{-}} &
  \multicolumn{1}{c|}{+20.20} &
  - &
  \multicolumn{1}{c|}{-} &
  \multicolumn{1}{c|}{-} &
  \multicolumn{1}{c|}{\textbf{\begin{tabular}[c]{@{}c@{}}31.68\\ ± \\ 6.4\end{tabular}}} &
  \textbf{-} \\ \hline
\end{tabular}%
}
\caption{Comparing the macro F1-Score of GenRDK and our **Best Approach (Entity Mention Description + Entity Mention Hypernym + Entity Mention Type + Dynamic Weighted Score). The DocRED dataset provides a blind test set, which prevented us from directly evaluating the performance of the test set. Result for GenRDK taken from \citep{10.1145/3589334.3645678}}
\label{tab:comparison_with_benchmark}
\end{table*}

\paragraph{Impact of Dynamic Weighted Score} \mbox{}\\
\label{para:impact_of_dynamic_weighted_score}
After observing a decrease in performance, we incorporated Weighted Score and Confidence into Equation \ref{eq:weighted_score_formula}. This significantly improved macro F1-Score. Too many similarity scores overshadowed relevant features, so higher weight was assigned to Entity Mention Descriptions due to their importance, while other side information received equal weight.

However, assigning weights alone didn’t fully address discrepancies from multiple side information sources. To tackle this, a Confidence Score was introduced, calculated using the mean and standard deviation of similarity scores. A high mean indicates strong alignment, while a low standard deviation ensures agreement among scores, enhancing reliability. For each entity pair, the highest similarity score determines the correct unseen relation label. Inconsistent measures (high standard deviation) highlight noise, and the confidence score mitigates this, improving prediction reliability. As shown in Table \ref{tab:ablation_study_full_unseen_relation_inference}, the Best Approach (Entity Mention Description + Hypernym + Type + Dynamic Weighted Score) outperformed the approach without dynamic weighting by 20.62\% (RE-DocRED), 24\% (DocRED), and 14.84\% (MEN-Dataset). This demonstrates the effectiveness of incorporating contextual information and dynamic weighting.

\subsection{Comparing with Benchmark Approach}
\label{ssec:comparingwithbenchmarkapproach}

In Section \ref{ssec:ablationstudy}, we compared different approaches to highlight the impact of Entity Side Information on predicting unseen relation labels. To assess relative performance, we compared our Best Approach (Entity Mention Description + Hypernym + Type + Dynamic Weighted Score) with GenRDK, the only existing Document-Level ZSRE method \citep{10.1145/3589334.3645678}. Table \ref{tab:comparison_with_benchmark} shows the comparison based on macro F1-Score.

Our Best Approach outperformed GenRDK, with an average improvement of 40.04\% for \( n \in \{10\} \) in RE-DocRED and DocRED. However, for \( n \in \{5\} \), the improvement was smaller (12.83\%), and no comparison was possible for \( n \in \{15\} \) due to missing GenRDK results. This suggests our approach excels with larger unseen label groups but shows diminishing gains with fewer labels. For 
\( n \in \{5\} \), our approach had a lower variance than GenRDK, indicating more stable predictions. However, for \( n \in \{10\} \), while achieving a higher macro F1-Score, our approach exhibited higher variance, suggesting sensitivity to specific relation labels. Detailed analysis revealed that one relation label group (r=3) had significantly lower F1-Scores and higher variance, likely due to fewer relation instances, semantic ambiguity, or errors in side information (e.g., entity types or hypernyms). This highlights the need for further investigation into label complexity and data quality.

\subsection{Exploring Inter and Intra-Sentential Relation Extraction Capabilities}
\label{sec:interandintrarelationextraction}
The proposed Document-Level ZSRE uniquely uses Entity Mention Descriptions instead of entire documents to predict relations. These descriptions capture the context of entity mentions from the original news article, with the head and tail entity descriptions concatenated as input. To ensure this approach doesn’t impact the performance, results were analyzed by comparing MEN-Dataset and RE-DocRED using relation label lists \( n \in \{5, 10\} \). Since RE-DocRED includes the full dataset (unlike DocRED, which only has 20\%), it was used in the experiments.

In Appendix \ref{sec:sentence_gaps_between_head_and_tail}, we present the results of an experiment conducted to evaluate the effectiveness of DocZSRE-SI in performing both inter-sentential and intra-sentential ZSRE. For both datasets (MEN-Dataset and RE-DocRED), the highest accuracy is observed when the sentence gap is 0, indicating that intra-sentence relations are easier to identify. Accuracy drops as the sentence gap increases in both datasets, suggesting that inter-sentence relations are harder to capture, for the MEN-Dataset at \( n \in \{5\} \), the accuracy declines from 56.69\% (gap = 1) to 16.93\% (gap $\geq$ 5). Meanwhile for RE-DocRED at \( n \in \{5\} \), the accuracy drops from 54.02\% (gap = 1) to 46.69\% (gap $\geq$ 5). Incorrect predictions increase significantly for larger sentence gaps. 

The results clearly show that sentence gaps negatively affect relation extraction performance. This is especially pronounced for inter-sentence relations (gap $\geq$ 1), where accuracy consistently declines as the gap widens. For \( n \in \{10\} \), the percentage of correct is lower than incorrect, as discussed in the previous section. This analysis demonstrates that while intra-sentence relations are well-handled by the proposed solution, inter-sentence relations remain a significant challenge when the gaps increase. Addressing this gap will require advanced modelling techniques, improved datasets, and leveraging external knowledge sources.

\section{Conclusion}
\label{sec:conclusion}
This paper introduces the Document-Level Zero-Shot Relation Extraction with Entity Side Information (DocZSRE-SI) framework, designed to predict unseen relation labels, especially in dynamic scenarios with emerging relations. A core component is Entity Side Information, which generates features like Entity Mention Descriptions, Types, and Hypernyms to provide contextual details for accurate predictions. Using Entity Side Information instead of the full document allows the model to focus on meaningful features of entities, reducing noise and improving generalization to unseen relations. Evaluations on standard datasets, such as DocRED and RE-DocRED, and language-specific datasets, like MEN-Dataset, demonstrate the framework’s adaptability. Although the Best Approach shows promising results, the high variability in performance compared to the benchmark is worth further investigation in the future. Despite this, the work presented in this chapter provides a starting point for improving Document-Level ZSRE methods for real-world use. For future work, we plan to enhance the Zero-Shot Relation Extraction Module to reduce variance and expand the application of Entity Mention Side Information to Supervised DocRE.

\section{Limitations}
\label{sec:limitations}
While our proposed framework, DocZSRE-SI, outperforms existing Document-Level ZSRE, it shows higher variance in certain scenarios, such as when \( n \in \{10\} \). Our investigation revealed that the approach is sensitive to semantically similar relation labels. Additionally, relying on entity types introduces ambiguity when using generic types like \texttt{MISC}, highlighting the need for more specific and accurate entity type assignments. Finally, the framework is simpler than some LLM-based approaches, but it involves multiple Similarity Score calculations and Dynamic Weighted Scores, which are computationally intensive for large-scale applications. We also acknowledge the concern about relying on manually chosen parameters or coefficients that may require adjustment for different applications or use cases. However, in our approach, the coefficients used for combining similarity scores are fixed and are selected after a few iterations of preliminary experiments on a validation set, rather than being tuned for each dataset or domain.

\section{Acknowledgement}
\label{sec:acknowledge}
We would like to acknowledge the responsible use of Generative AI tools, which assisted in error checking and improving the clarity of my writing in compliance with academic integrity guidelines. Large Language Models (LLMs) like gpt-4o-mini were employed to generate Entity Mention Descriptions and Hypernyms. Part of this project was funded by the Malaysian Fundamental Research Grant Scheme (FRGS) FRGS/1/2022/ICT02/MUSM/02/2.

\bibliography{custom}
\appendix

\section{Example of Generated Entity Mention Description}
\label{sec:entity_mention_description_appendix}

Consider the following news article snippet as an example: 

\textit{KUALA LUMPUR: Malaysia logged another 21,072 Covid-19 infections as of noon today.  Health director-general Tan Sri Dr Noor Hisham Abdullah said of the fresh cases recorded today, 86 (or 0.41 per cent) were in Categories 3, 4 and 5.  In Category 5, there were eight cases (0.04 per cent); Category 4 with 20 cases (0.09 per cent), and Category 3 with 58 (0.28per cent) ...}. 

To generate a description for Tan Sri Dr. Noor Hisham Abdullah (PERSON), the content will be generated based on the entity mention. The generated description is: 

\textit{Tan Sri Dr Noor Hisham Abdullah is the Health Director-General of Malaysia, responsible for reporting and managing the country's Covid-19 response during the pandemic}. 

This description provides contextual information about the entity, helping to enhance the model's understanding of its role and significance within the document.

\section{Prompt Template for Role-Based Embeddings}
\label{sec:prompt_role_based_embedding}
\begin{itemize}
    \item Head: \texttt{\{head\_type\} acting as a subject, described as \{head\_hypernym\}}
    \item Tail: \texttt{\{tail\_type\} acting as a subject, described as \{tail\_hypernym\}}
\end{itemize}

\section{Prompt Template for Context Embeddings}
\label{sec:prompt_context_embedding}
\begin{itemize}
    \item \texttt{Relation between \{head\_hypernym\} and \{tail\_hypernym\}}. 
\end{itemize}
    
\section{Result of Experiment to Evaluate the Inter and Intra-Sentential Capabilities of DocZSRE-SI}
\label{sec:sentence_gaps_between_head_and_tail}

Table \ref{tab:sentence_gaps_between_head_and_tail} shows the result of our analysis to understand the effectiveness of our approach when handling both inter and intra-sentential relation extraction.

\begin{table*}[ht!]
\centering
\resizebox{\textwidth}{!}{%
\begin{tabular}{|c|cccccc|cccccc|}
\hline
 &
  \multicolumn{6}{c|}{MEN-Dataset} &
  \multicolumn{6}{c|}{RE-DoCRED} \\ \hline
 &
  \multicolumn{3}{c|}{5} &
  \multicolumn{3}{c|}{10} &
  \multicolumn{3}{c|}{5} &
  \multicolumn{3}{c|}{10} \\ \hline
\begin{tabular}[c]{@{}c@{}}Sentence\\ Gap (H-T)\end{tabular} &
  \multicolumn{1}{c|}{\begin{tabular}[c]{@{}c@{}}Total\\ Instance\end{tabular}} &
  \multicolumn{1}{c|}{\begin{tabular}[c]{@{}c@{}}Correct\\ (\%)\end{tabular}} &
  \multicolumn{1}{c|}{\begin{tabular}[c]{@{}c@{}}Incorrect\\ (\%)\end{tabular}} &
  \multicolumn{1}{c|}{\begin{tabular}[c]{@{}c@{}}Total\\ Instance\end{tabular}} &
  \multicolumn{1}{c|}{\begin{tabular}[c]{@{}c@{}}Correct\\ (\%)\end{tabular}} &
  \begin{tabular}[c]{@{}c@{}}Incorrect\\ (\%)\end{tabular} &
  \multicolumn{1}{c|}{\begin{tabular}[c]{@{}c@{}}Total\\ Instance\end{tabular}} &
  \multicolumn{1}{c|}{\begin{tabular}[c]{@{}c@{}}Correct\\ (\%)\end{tabular}} &
  \multicolumn{1}{c|}{\begin{tabular}[c]{@{}c@{}}Incorrect\\ (\%)\end{tabular}} &
  \multicolumn{1}{c|}{\begin{tabular}[c]{@{}c@{}}Total\\ Instance\end{tabular}} &
  \multicolumn{1}{c|}{\begin{tabular}[c]{@{}c@{}}Correct\\ (\%)\end{tabular}} &
  \begin{tabular}[c]{@{}c@{}}Incorrect\\ (\%)\end{tabular} \\ \hline
0* &
  \multicolumn{1}{c|}{465} &
  \multicolumn{1}{c|}{50.65} &
  \multicolumn{1}{c|}{49.35} &
  \multicolumn{1}{c|}{1463} &
  \multicolumn{1}{c|}{57.5} &
  42.5 &
  \multicolumn{1}{c|}{3277} &
  \multicolumn{1}{c|}{59.05} &
  \multicolumn{1}{c|}{40.95} &
  \multicolumn{1}{c|}{5135} &
  \multicolumn{1}{c|}{32.99} &
  67.01 \\ \hline
1 &
  \multicolumn{1}{c|}{356} &
  \multicolumn{1}{c|}{56.69} &
  \multicolumn{1}{c|}{43.31} &
  \multicolumn{1}{c|}{795} &
  \multicolumn{1}{c|}{40.5} &
  59.5 &
  \multicolumn{1}{c|}{1251} &
  \multicolumn{1}{c|}{54.02} &
  \multicolumn{1}{c|}{45.98} &
  \multicolumn{1}{c|}{3043} &
  \multicolumn{1}{c|}{30.89} &
  69.11 \\ \hline
2 &
  \multicolumn{1}{c|}{206} &
  \multicolumn{1}{c|}{45.22} &
  \multicolumn{1}{c|}{54.78} &
  \multicolumn{1}{c|}{495} &
  \multicolumn{1}{c|}{26.26} &
  73.74 &
  \multicolumn{1}{c|}{548} &
  \multicolumn{1}{c|}{50.55} &
  \multicolumn{1}{c|}{49.45} &
  \multicolumn{1}{c|}{711} &
  \multicolumn{1}{c|}{39.24} &
  60.76 \\ \hline
3 &
  \multicolumn{1}{c|}{54} &
  \multicolumn{1}{c|}{41.11} &
  \multicolumn{1}{c|}{58.89} &
  \multicolumn{1}{c|}{132} &
  \multicolumn{1}{c|}{37.58} &
  62.42 &
  \multicolumn{1}{c|}{405} &
  \multicolumn{1}{c|}{49.14} &
  \multicolumn{1}{c|}{50.86} &
  \multicolumn{1}{c|}{459} &
  \multicolumn{1}{c|}{37.69} &
  62.31 \\ \hline
4 &
  \multicolumn{1}{c|}{32} &
  \multicolumn{1}{c|}{30.38} &
  \multicolumn{1}{c|}{69.62} &
  \multicolumn{1}{c|}{105} &
  \multicolumn{1}{c|}{30.14} &
  69.86 &
  \multicolumn{1}{c|}{373} &
  \multicolumn{1}{c|}{48.26} &
  \multicolumn{1}{c|}{51.74} &
  \multicolumn{1}{c|}{397} &
  \multicolumn{1}{c|}{31.49} &
  68.51 \\ \hline
$\geq$ 5 &
  \multicolumn{1}{c|}{189} &
  \multicolumn{1}{c|}{16.93} &
  \multicolumn{1}{c|}{83.07} &
  \multicolumn{1}{c|}{485} &
  \multicolumn{1}{c|}{11.75} &
  88.25 &
  \multicolumn{1}{c|}{829} &
  \multicolumn{1}{c|}{46.69} &
  \multicolumn{1}{c|}{53.31} &
  \multicolumn{1}{c|}{976} &
  \multicolumn{1}{c|}{30.74} &
  69.26 \\ \hline
\end{tabular}%
}
\caption{The overall count of sentence gaps between head and tail entities and the percentage of correct and incorrect predictions for each gap category (0, 1, 2, 3, 4, and $\geq$ 5). *0 refers to condition where Head Entity and Tail Entity are in same sentence (Intra-Sentential RE).}
\label{tab:sentence_gaps_between_head_and_tail}
\end{table*}

\end{document}